\documentclass[10pt,twocolumn,letterpaper]{article}
\usepackage{acpr}

\usepackage{cite}
\usepackage{times}
\usepackage{epsfig}
\usepackage{graphicx}
\usepackage{amsmath}
\usepackage{amssymb}
\usepackage{tablefootnote}
\usepackage{color}
\usepackage{subcaption}
\usepackage{multirow}
\usepackage{caption}
\usepackage[pagebackref=true,breaklinks=true,letterpaper=true,colorlinks,bookmarks=false]{hyperref}

\acprfinalcopy 
\ifacprfinal\fi

\begin{document}

\title{Image Distortion Detection using Convolutional Neural Network}
\author{
Namhyuk Ahn\\
Ajou University\\
Suwon, Korea\\
{\tt\small aa0dfg@ajou.ac.kr}
\and
Byungkon Kang\\
Ajou University\\
Suwon, Korea\\
{\tt\small byungkon@ajou.ac.kr}
\and
Kyung-Ah Sohn\\
Ajou University\\
Suwon, Korea\\
{\tt\small kasohn@ajou.ac.kr}
}

\maketitle
\begin{abstract}
Image distortion classification and detection is an important task in many applications. For example when compressing images, if we know the exact location of the distortion, then it is possible to re-compress images by adjusting the local compression level dynamically.  In this paper, we address the problem of detecting the distortion region and classifying the distortion type of a given image. We show that our model significantly outperforms the state-of-the-art distortion classifier, and report accurate detection results for the first time. We expect that such results prove the usefulness of our approach in many potential applications such as image compression or distortion restoration.
\end{abstract}

\section{Introduction}
\label{sec:introduction}
With the development of the Internet and mobile devices, demand for streaming media and cloud service have skyrocketed.
These services need a lot of storage to store multimedia, and it is crucial to compress data using lossy compression techniques before storing. Higher compression level is better in terms of storage, but it could cause serious local distortion to images. However, if we can detect the region in which the distortions occurred, we can correct the problem by dynamic compression techniques. Such techniques reduce the compression level of detected distortion regions and re-compress using the reduced level.

\begin{figure*}[ht]
	\centering
	\begin{subfigure}[b]{0.24\textwidth}
		\includegraphics[width=\textwidth, height=\textwidth]{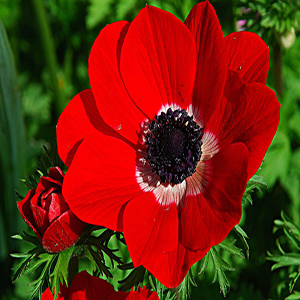}
		\caption{Reference}
		\label{fig:dataset-ref}
	\end{subfigure}
	\begin{subfigure}[b]{0.24\textwidth}
		\includegraphics[width=\textwidth, height=\textwidth]{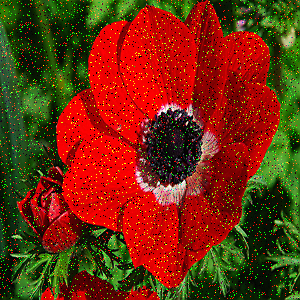}
		\caption{\textit{Classification}}
		\label{fig:dataset-task1}
	\end{subfigure}
	\begin{subfigure}[b]{0.24\textwidth}
		\includegraphics[width=\textwidth, height=\textwidth]{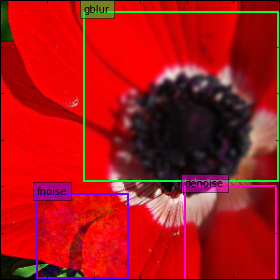}
		\caption{\textit{Detection-basic}}
		\label{fig:dataset-task2}
	\end{subfigure}
	\begin{subfigure}[b]{0.24\textwidth}
		\includegraphics[width=\textwidth, height=\textwidth]{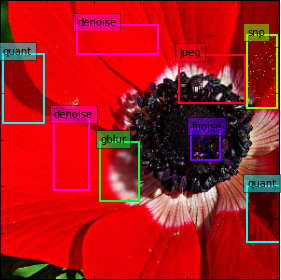}
		\caption{\textit{Detection-difficult}}
		\label{fig:dataset-task2-difficult}
	\end{subfigure}
	\caption{Example of \textit{Flickr-Distortion} dataset. (a) is the reference image,
	(b) is the distorted image with salt and pepper noise for classification task.
	(c) and (d) are both for detection task, with different levels of detection
	difficulty described in Section~\ref{sec:dataset-task2}.}
	\label{fig:dataset}
\end{figure*}

Our motivation for conducting this research is to build a system that
detects the distortion region and performs compression dynamically.
Hence, automatic distortion detection is an essential part of this system. However, despite the importance of this task, recent image quality assessment (IQA) methods only focus on predicting perceptual quality scores, such as the  mean opinion score (MOS) \cite{kang2014, bosse2016, bianco2016, kang2015}.

One might question the validity of the assumption that multiple distortions exist in a single image. While it is true that distortions in an image are likely to occur globally rather than locally,  we consider a situation where individual images are assembled to form a larger one.  For example, when creating a panorama picture, the compression of each shot may be subjected to independent distortion. When the individual shots are combined to form the full photo, it might end up with localized distortions.

In recent years, many deep learning based IQA approaches have been proposed, especially for non-reference IQA (NR-IQA). NR-IQA methods perform image quality assessment without any direct comparison between the reference and the distorted image. The IQA-CNN~\cite{kang2014} model is the first such model that applies deep learning in IQA task. This model uses a convolutional neural network (CNN) composed of five layers, that achieves the result comparable to the full-reference IQA (FR-IQA) methods, such as FSIM~\cite{zhang2011}.

Deeper networks have been used in~\cite{bosse2016}, whose structure is inspired by the VGGNet~\cite{simonyan2014}, and yields results that surpass FR-IQA approaches. DeepBIQ~\cite{bianco2016} is the first to use pre-trained CNN and show state-of-the-art result in the IQA task. However, although there exist many outstanding results, the distortion classification task remains mostly unchallenged. Two notable exceptions to this are the IQA-CNN+ and the IQA-CNN++~\cite{kang2015}, that predict both the MOS and the distortion type with the similar network used in IQA-CNN. One shortcoming of these models is that they are shallow architectures, and thus might have limited capacity to successfully solve our task.

In addition, to best to our knowledge, distortion detection task with deep learning method has not been applied yet. The reason is twofold: First, there is no sufficiently large distortion detection dataset suitable for deep learning, and second, detection task is a much more challenging problem than a classification or predicting MOS are. The difficulty of distortion detection is based on the fact that images can have heterogeneous and multiple distortion types. In general, most IQA datasets contain only homogeneous distortion types that make prediction relatively easy.

To tackle these issues, we created a new dataset for both distortion
classification and detection. Then we apply pre-trained CNNs such as VGGNet\cite{simonyan2014} and ResNet\cite{he2016} for distortion classification. Finally, we use deep learning based detection methods such as single shot multi-box detector (SSD)\cite{liu2015} to locate the distortion regions.

Our main contributions are as follows: 1) We create a dataset for distortion classification and detection task. There are no publicly available such datasets. 2) We fine-tune a pre-trained CNN to this dataset to get high performance. 3) We propose a distortion detection system that uses an existing CNN model trained to achieve good performance in object detection task. To the best of our knowledge, our method is the first attempt to use deep learning
based detection method in distortion detection task.

Section~\ref{sec:dataset} presents our dataset, followed by the description for our system in Section~\ref{sec:method}. Experiments and results are explained in Section~\ref{sec:experiment}, and finally
conclusion in Section~\ref{sec:conclusion}.

\section{CNN for Distortion Recognition}
\label{sec:method}
The main model we choose for our distortion recognition system is a
convolutional neural network (CNN)~\cite{cnn}. CNNs have been widely used and verified over
a variety of image understanding tasks~\cite{cnn_charrec,cnn_imgseg,alexnet}.
The overall structure we use is a `Y'-shaped
CNN that performs distortion classification and detection simultaneously. Both the
classifier and the detector share the same feature-extraction portion, after which the
structure splits into two sets of layers to perform classification and detection, respectively.
\subsection{Distortion Classification}
In this paper, we experiment with VGG-16\cite{simonyan2014} and
ResNet-101\cite{he2016} models. Both networks are variants of CNN which
consist of several convolution, pooling and fully-connected (FC) layers to
recognize images. VGG-16 has 13 convolution layers and 3 FC layers.
Because of the simplicity of this network, many researchers use the VGG-16
as a base network. The Atrous VGGNet is introduced by
DeepLab\cite{chen2016} with an architecture similar to the VGGNet, but with a
difference in the number of parameters in the final fully-connected layers, and
its use of Atrous convolution that allows for the processing of arbitrary-sized
field-of-views.

ResNet uses \textit{residual connections} to avoid the degradation problem.
Without residual connections, deep networks are known to not only overfit
but also show increasing training error.
Unlike the VGGNet, ResNet-101 uses 101 layers with only the last
layer being fully-connected. Additionally, a global average pooling
technique is used to reduce the number of parameters.

In practice, training the entire CNN from scratch is a difficult and
time-consuming job. Also, if the dataset does not have sufficient
training data, training does not converge well. Therefore, it is common
to use pre-trained networks which have been trained on large
external datasets such as ImageNet\cite{deng2009}. This transfer learning
strategy works well if the distribution of the source dataset (used for
pre-trained) and target dataset are similar. As ImageNet and our dataset have
similar distribution, we use CNNs pre-trained on ImageNet for all our experiments.
\subsection{Distortion Detection}
For the distortion detection task, we use the single shot multibox detector
(SSD)\cite{liu2015}. With the development of CNN, many detection methods
have been proposed such as R-CNN\cite{girshick14}, Faster R-CNN\cite{ren2015},
YOLO\cite{redmon2016}, OverFeat\cite{overfeat}, and SSD\cite{liu2015}. R-CNN and
its variants perform state-of-the-art detection, while inference time is very slow due to the
limitation of their architecture. On the other hand, YOLO and SSD are
real-time detection algorithms, with SSD outperforming YOLO.

SSD computes multi-scale feature maps for detection by adding extra convolution
layers at the end of the base network. Then six output feature maps from
different convolution layers are concatenated to form the final layer.
With this idea, SSD effectively detects from objects of various sizes using a
single, simple architecture, since the output maps from lower layer tend to
capture fine-grained details of object. Predictors of SSD rely on
convolution layers instead of the conventional fully-connected layers,
to reduce inference time.

In our experiments, we use the best setting for SSD:
1) Use Atrous VGG-16 as a baseline network since it shows similar result
with faster running time.
2) We use 300x300 as the input dimension. If we increase input dimension
to 500x500, the inference becomes much slower while the performance gain is relatively small.
This shows that using 300x300 input can capture small-sized objects reasonably well in short time via multi-scale feature maps.
3) On the contrary to the SSD used in object detection task, the only data
augmentation we use is the horizontal flip. This is because affine transformations
might corrupt the details of the distortion, such as in the case of scaling
and shearing.

\section{Flickr-Distortion Dataset}
\label{sec:dataset}
We creat a new dataset named \textit{Flickr-Distortion} dataset to
evaluate image distortion classification and detection task.
To make this dataset, we first collect 804 reference images from Flickr
with similar way to NUS-WIDE\cite{chua2009} dataset, and make distorted images using
the collected reference images. We use eight distortion types:
1) Gaussian white noise (GWN), 2) Gaussian blur (GB), 3) salt and pepper noise (S\&P),
4) quantization noise, 5) JPEG compression noise, 6) low-pass noise,
7) denoising and 8) fnoise.

The reason we do not use the LIVE dataset directly is that
it contains \textit{global} distortions, whereas we deal with \textit{local} distortions.
Furthermore, prevalent distortion dataset such as LIVE\cite{sheikh2005} or TID2013\cite{ponomarenko2015}
have insufficient amount of reference images which are not suitable for training deep learning models.

\subsection{Flickr-Distortion-Classification Dataset}
In the dataset for distortion classification task, each reference image
is distorted using eight distortion types with three levels.
Thus, a single reference image results in 24 distorted images.
The distortion procedure follows the LIVE dataset\cite{sheikh2005},
and includes such distortion types as homogeneous distortion.
The distortion is applied to the entire image (see Fig. \ref{fig:dataset-task1}).
We create 19,296 distorted image in total, and randomly split data into 60\%
training, 20\% validation and 20\% test set.

We implemented the generation of most noises except fnoise, for which we used
the scikit-image\cite{van2014} python library. Detailed values we use are as
follows: Gaussian noise is generated with three values of variances:
\{0.0125, 0.025, 0.05\}, and the amount of salt and pepper noise are same as
Gaussian noise. For the Gaussian blur, we use the three sigma levels
\{1.5, 3, 6\} and in JPEG compression we use \{20, 10, 5\} quality levels.
To implement the low-pass noise, we simply scaled images with ratios
\{0.3, 0.1, 0.03\}, and re-sized to the original image size.
We implemented denoising with non-local means algorithm\cite{buades2005} using
factors \{0.04, 0.06, 0.08\}, but with fixed batch size and patch distance of
7 and 11, respectively. Finally, fnoise is implemented using noise level
$1/f$ with factor $f\in\{2.5, 5, 10\}$.

\subsection{Flickr-Distortion-Detection Dataset}
\label{sec:dataset-task2}
Unlike the \textit{Classification} dataset, each image in the detection
dataset can have heterogeneous and multiple distortions, as shown in
Fig.~\ref{fig:dataset-task2}. Since SSD network used in the detection task
accepts images of dimension 300x300 as input, we crop the center of correct
size before applying distortion.

Distortion levels are chosen uniformly at random with range of minimum and maximum
values used in the \textit{Classification} dataset. When choosing the
distortion regions, we sample the number of regions in a single image from
a uniform distribution over the interval [1, 4]. The ratio of region size to
image size is also picked uniformly at random from [0.3, 0.7] (average is 0.43).
We generate 20 images per reference image, with a total of 16,080 images created.
For evaluation, we split data into 80\% training and 20\% for test.

The assumption on the number of regions and sizes in the \textit{Detection}
dataset is quite reasonable. However in practice, there may be many small
regions of distortions. Therefore, as in Fig. \ref{fig:dataset-task2-difficult},
we created another dataset named \textit{Detection-difficult} sets
(Above dataset is named with \textit{Detection-basic}).
In this dataset, the minimum and maximum number of regions per image are 5 and 9,
respectively. The ratio of region and image size is changed to 0.1 and 0.3,
with an average of 0.18.
\section{Experiment}
\label{sec:experiment}
In this section, we describe our experimental results. For ease of exposition, we separate
the report on classification and detection into subsections. With the exception of
pre-training, our dataset is given in Section~\ref{sec:dataset}.
\subsection{Distortion Classification}
We first evaluate the distortion classification task using pre-trained networks.
To do this, we remove the last fully-connected layer in the pre-trained network
and freeze all layers in network. Then, we add a new fully-connected layer
suited for the number of classes of our dataset.
Only this new layer is trained from scratch. Since our classification dataset
homogeneously distorted as in LIVE~\cite{sheikh2005}, we center-crop the
images so the size fit the input layer of the network without concern for
equal distribution of the distortion.
We also evaluate a fine-tuned network. The training procedure of the fine-tuned
network is the same as that of the non-fine-tune version, but in this case we
let the gradient propagate through all layers.
\begin{table}[h]
\begin{center}
\begin{tabular}{|c | c | c | c|}
\hline
Method        & w/o finetuning & w/ finetuning    & FPS
\\\hline
IQA-CNN+      & -        & 0.820          & 166
\\
IQA-CNN++     & -          & 0.782          & \textbf{250}
\\
\hline
VGG-16        & 0.858        & 0.984          & 83
\\
Atrous VGG-16 & 0.855        & 0.984          & 90
\\
ResNet-101    & 0.926        & \textbf{0.988} & 20
\\ \hline
\end{tabular}
\end{center}
\caption{Quantitative results of classification accuracy and inference speed in
distortion classification task. VGG- and ResNet-based models outperform IQA-CNN variants.
}
\label{tb:task1-result}
\end{table}

The result of the classification task is given in Table~\ref{tb:task1-result}.
To verify what effect pre-training has, we use IQA-CNN+ and
IQA-CNN++\footnote{We re-implemented these networks using
TensorFlow\cite{abadi2016}. Note that we remove linear regression layer for predicting MOS.}~\cite{kang2015} as baseline.
As can be seen in the results, the pre-trained networks outperform
baseline networks.
Since all pre-trained networks have deeper architectures compared to the
baseline, they are suitable for complex data due to the high network capacity.
Moreover, the fine-tuning procedure makes the network better-adapt to the new data.

Among non fine-tuned pre-trained networks, ResNet outperforms the VGGNet family.
This is due to the output of VGGNet being 4096-dimensional, which is twice
as large as that of ResNet.

However, all three networks show similar performance after being fine-tuned.
We conjecture that this is probably because all networks have relatively
large enough capacity to handle this task.
Unlike the accuracy, inference time shows a large gap among different
architectures. ResNet is much slower than the VGGNet family,
and Atrous VGGNet is faster than the vanilla VGGNet, since Atrous VGGNet
subsamples parameters in its final two fully-connected layers.
\begin{figure}[t]
	\centering
	\begin{subfigure}[b]{0.23\textwidth}
		\includegraphics[width=\textwidth, height=\textwidth]{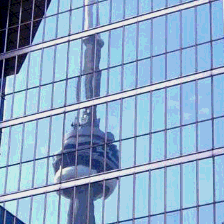}
		\caption{quantization$\rightarrow$fnoise}
		\label{fig:false2}
	\end{subfigure}
	\begin{subfigure}[b]{0.23\textwidth}
		\includegraphics[width=\textwidth, height=\textwidth]{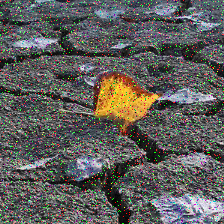}
		\caption{S\&P$\rightarrow$GWN}
		\label{fig:false4}
	\end{subfigure}
	\caption{Example of mis-classified images. (a) has quantization distortion but is
	predicted as fnoise. (b) Salt and pepper noise classified as Gaussian white noise.}
	\label{fig:falsecase}
\end{figure}

In most cases our model can classify distortion types very well.
However, as in Fig. \ref{fig:falsecase}, some images are commonly mis-classified.
For example, salt and pepper noise is often mistaken as Gaussian white noise as
seen in Fig. \ref{fig:false4} and vice versa.
Fig. \ref{fig:false2} shows a case where the image does not show abrupt change in color,
in which case the model also confuses quantization noise with fnoise.
Such problems can be alleviated if we directly compare the given image
with a reference image, but in practice, there are restrictions on using
reference images. Hence our approach well-balances between the accuracy and
practicality.
\subsection{Distortion Detection}
In this section, we present the results on distortion detection experiment
with SSD. Here, we only use the Atrous VGG-16 and ResNet-101 since VGG-16
and Atrous VGG-16 have similar performance but Atrous VGG-16 is faster.
The IQA-CNN+ achieves reasonable result with very fast inference time, however, since this model only has single convolution layer, it is not appropriate  to use the SSD that needs multiple convolution layers.

When training the network, we use the pre-trained CNN, which is fine-tuned
over the \textit{Classification} dataset, as a base network and stack SSD layer
on top of the base network.
In the evaluation step, we use the mean average precision (mAP) metric which
measures the average precision of each class when the intersection over union
(IoU) of the bounding box is one of \{0.5, 0.75, 0.9\}.
In typical object detection tasks, performance evaluation is usually done with IoU @0.5.
However, in our assumed scenario of finding the distortion regions for the
purpose of applying local dynamic compression, finding accurate boxes is vital.
This is why we use a variety of IoU thresholds to assess the degree of our algorithm's
region detection accuracy. Result of experiments are in Table \ref{tb:task2-result1}.
\begin{table}[h]
\begin{center}
\begin{tabular}{|c | c | c | c | c|}
\hline
\multirow{2}{*}{Method} & \multirow{2}{*}{FPS} & \multicolumn{3}{|c|}{mAP. IoU:}
\\\cline{3-5}
              &             & @ 0.5          & @ 0.75         & @ 0.9
\\\hline
ResNet-101    & 16          & 0.910          & 0.900          & 0.842
\\
Atrous VGG-16 & \textbf{63} & \textbf{0.919} & \textbf{0.906} & \textbf{0.873}
\\
\hline
\end{tabular}
\end{center}
\caption{Results of distortion detection (mAP) and inference speed over a variety of IoU
threshold values. Atrous VGG performs slightly better than ResNet with higher FPS.}
\label{tb:task2-result1}
\end{table}

Surprisingly, there does not seem to be any advantages of using ResNet
over VGGNet in this experiment.
We believe that this is because VGGNet's capacity is large enough to fit to the given data.
In addition, Atrous VGG-16 excels ResNet-101 in terms of inference time, and it can be done in
real-time on state-of-the-art GPUs such as Maxwell TITAN X.
\begin{table}[h]
\begin{center}
\begin{tabular}{|c | c | c | c|}
\hline
\multirow{2}{*}{Train data $\rightarrow$ Test data}& \multicolumn{3}{|c|}{mAP. IoU:} \\ \cline{2-4}
& @0.5 & @0.75 & @0.9 \\ \hline
\textit{basic}   $\rightarrow$ \textit{basic}     & 0.919 & 0.906 & 0.873 \\
\textit{difficult} $\rightarrow$ \textit{basic}    & 0.915 & 0.864 & 0.728 \\
\hline
\textit{basic}   $\rightarrow$ \textit{difficult}  & 0.717 & 0.467 & 0.109 \\
\textit{difficult} $\rightarrow$ \textit{difficult} & 0.908 & 0.895 & 0.785 \\
\hline
\end{tabular}
\end{center}
\caption{Transfer experiment results on detection task with Atrous VGG-16 network.}
\label{tb:task-result2}
\end{table}

\begin{figure}[t]
\begin{subfigure}[b]{0.236\textwidth}
    \includegraphics[width=\textwidth, height=\textwidth]{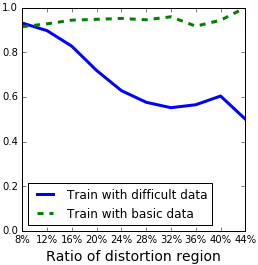}
    \caption{Test acc. with \textit{basic} data}
    \label{fig:basic_test}
\end{subfigure}
\begin{subfigure}[b]{0.236\textwidth}
    \includegraphics[width=\textwidth, height=\textwidth]{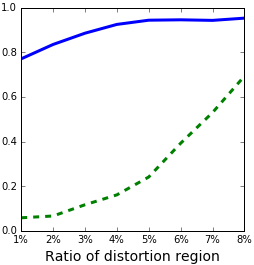}
    \caption{Test acc. with \textit{difficult} data}
    \label{fig:difficult_test}
\end{subfigure}
\caption{Relationship between distortion size and detection accuracy with IOU@0.9. (a) shows test accuracy with \textit{basic} data and (b) is for \textit{difficult} scenario.}
\label{fig:box-size}
\end{figure}

As described in Section \ref{sec:dataset-task2}, distortion may occur in small local regions
in real world. Therefore, evaluating using only the \textit{basic} dataset might not be
a desirable strategy.
To further investigate, we conducted a set of transfer learning experiments that evaluates
the four combinations of training-testing scenarios, where the training and testing datasets
can be either \textit{basic} or \textit{difficult}.
Table \ref{tb:task-result2} shows that the models trained and tested on the same type
of dataset yield the best performance.
This is natural since the model trained using only the \textit{basic} set cannot catch
distortions with small region, while training only on the \textit{difficult} set tends to
drive the detector towards finding small regions.
Note that in \textit{basic} $\rightarrow$ \textit{difficult}
case, the trained model performs poorly when the IoU threshold is large, since it misses
most small-size regions.
The graph in Fig.~\ref{fig:box-size} shows how the accuracy changes as the size of the
ground-truth regions changes. This also illustrates that training with \textit{basic} data
is good when the area of the region is large, but performs worse for small sizes
when trained on \textit{difficult} data.

To sum up, it is crucial to match the settings between train and test data.
But since we do not know much about the test set, it is better to train on the
\textit{difficult} set to better-cope with possible worst-case scenario based on the
assumption that distortion may occur in small local region.

\section{Conclusion}
\label{sec:conclusion}
We investigated the novel problem of classifying and detecting distortion in an image without reference image using CNN architectures. To do that, we created a new \textit{Flickr-Distortion} dataset to train on. In distortion classification, we used fine-tuned models that have been pre-trained on the ImageNet data, in order to reach convergence quickly. By doing so, we discovered that fine-tuned CNNs outperform other baseline models such as IQA-CNN+.
Furthermore, we experimented the distortion detection task with the SSD\cite{liu2015} model, which has not been addressed previously. We found that our architecture is able to efficiently classify and detect various distortions, despite using a single set
of weights for all distortion types.

We expect that our approach proves the usefulness of distortion detection in many applications such as dynamic compression technique or image reconstruction. One of our main discoveries is that the difference in the quality of training images and the testing images significantly affects the overall performance. This is not necessarily a surprising fact from a machine learning point-of-view, but it does have important practical implications. Since we do not know in advance the quality of the image we process, it might be difficult to guarantee the performance of our final system. We propose that we deploy our system after training on image sets consisting mostly of \textit{difficult} images, in order to cope with the worst-case-scenario.

As a future work, we are planning to further develop our system to handle multiple distortions in a specialized manner. Our current system tries to classify and detect multiple distortions using a single structure. To account for multiple distortions, one must have a high-capacity system that could potentially lead to overfitting. If we could devise an ensemble-like system that specializes for each distortion type, the system might be able to focus on quality-neutral generalization within each distortion.

\section*{Acknowledgement}
N.Ahn and K.-A. Sohn were supported by Basic Science Research Program through the National Research Foundation of Korea (NRF) funded by the Ministry of Education [NRF-2016R1D1A1B03933875], and B.Kang by [NRF-2016R1A6A3A11932796].

\bibliographystyle{ieee}
\bibliography{}
\end{document}